\documentclass{article}
\usepackage{spconf,amsmath,graphicx}

\usepackage{amssymb}
\usepackage[colorlinks,linkcolor=black,urlcolor=black,citecolor=black]{hyperref}
\usepackage{multirow}
\usepackage{subfigure}
\usepackage{enumitem}
\usepackage{tabularx}
\usepackage{amsmath}
\usepackage{booktabs}
\usepackage[capitalize]{cleveref}
\crefname{section}{Sec.}{Secs.}
\Crefname{section}{Section}{Sections}
\crefname{table}{Tab.}{Tabs.}
\Crefname{table}{Table}{Tables}
\crefname{figure}{Fig.}{Figs.}
\Crefname{figure}{Figure}{Figures}
\crefname{equation}{Eq.}{Eqs.}
\Crefname{equation}{Equation.}{Equations.}
\usepackage{lineno}

\title{LiCAF: LiDAR-Camera Asymmetric Fusion for Gait Recognition}
\name{Yunze~Deng, Haijun~Xiong, Bin~Feng
\thanks{This work was supported by the National Key R\&D Program of China (2023YFF0905401).}
\thanks{Corresponding author: Bin Feng (fengbin@hust.edu.cn.)}}

\address{Huazhong University of Science and Technology, Wuhan, China}

\begin{document}
%
\maketitle
\begin{abstract}
Gait recognition is a biometric technology that identifies individuals by using walking patterns. Due to the significant achievements of multimodal fusion in gait recognition, we consider employing LiDAR-camera fusion to obtain robust gait representations. However, existing methods often overlook intrinsic characteristics of modalities, and lack fine-grained fusion and temporal modeling. In this paper, we introduce a novel modality-sensitive network \textbf{LiCAF}  for LiDAR-camera fusion, which employs an asymmetric modeling strategy. Specifically, we propose Asymmetric Cross-modal Channel Attention (\textbf{ACCA}) and Interlaced Cross-modal Temporal Modeling (\textbf{ICTM}) for cross-modal valuable channel information selection and powerful temporal modeling. Our method achieves state-of-the-art performance (93.9\% in Rank-1 and 98.8\% in Rank-5) on the SUSTech1K dataset, demonstrating its effectiveness.
\end{abstract}
%


\section{Introduction}
\label{sec: Introduction}

Gait recognition technology typically extracts discriminative gait representations from the collected pedestrian gait data to identify distinct individuals, characterized by its non-contact and anti-camouflage nature. Nowadays, gait recognition technology has been widely applied in various fields, \textit{e.g.} public security, medical diagnosis, and sports science \cite{shen2022comprehensive, sheng2023data, cosma2024psymo, ma2023fine}.

Currently, most gait recognition methods mainly focus on RGB modality gait data. 
With the development of data collection devices and sensors, the strategies for collecting gait data become more diverse.
Therefore, many multi-modal gait recognition datasets and methods have emerged these years \cite{zheng2022gait, shen2023lidargait, han2022licamgait, cui2023multi, peng2023learning}, encompassing modalities such as RGB images, point clouds, and skeletons. 
Different modalities often carry unique and distinct gait information, hence many researchers exploit their complementary nature, utilizing the fused information to improve performance and address significant challenges in gait recognition, such as cloth-changing, occlusion, and complex outdoor environments \cite{shen2023lidargait, cui2023multi}. 

\begin{figure*}[htp]
    \centering
    \includegraphics[width=1\linewidth]{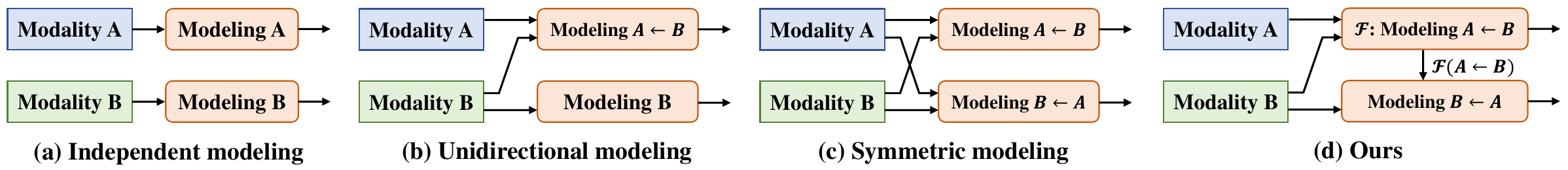}
    \caption{A simplified diagram of four commonly used strategies to model multi-modal information. In this context, Modeling $A\gets B$ indicates the process of modeling Modality A, incorporating supplementary information from Modality B. The notation $\mathcal{F}(A\gets B)$ represents the modeling results of Modality A, including information from Modality B. 
    }
    \label{fig: intro}
    \vspace{-0.5cm}
\end{figure*}

Typically, for multimodal gait recognition, LiDAR data can provide 3D geometric information and view-independent depth information of different body parts, while camera data can offer a silhouette of the human body \cite{han2022licamgait}. Therefore, complementary modeling of both modalities can fully capitalize on their strengths, providing robust and discriminative gait representations. However, most LiDAR-camera fusion methods in other fields such as autonomous driving \cite{bai2022transfusion, li2022bevformer} employ coarse-grained modeling of LiDAR and camera features, whereas gait recognition necessitates fine-grained modeling of gait features. Additionally, much contextual and motion information is lost due to the lack of temporal modeling, making these methods unsuitable for gait recognition. 

Another problem with current LiDAR-camera fusion methods is neglecting inherent characteristics between two modalities, lacking modality-sensitivity. Besides, the interaction of information between modalities during the modeling process is insufficient. 
Generally, features from different modalities locate in distinct subspaces and vary in the richness and complexity of the information they provide \cite{youpeng2021amvae}. Therefore, the contribution of these modalities to a cross-modal modeling process is not uniform, which means that if $A\gets B$ represents the supplementation of information from B into A during the modeling process, then $A\gets B$ is not equivalent to $B\gets A$ \cite{wang2020learning}. 
According to this, for different modeling processes that serve distinct functions, it is particularly important to carefully design the order of information supplementation and guidance between modalities, making the model more modality-sensitive. This order enables the model to fully understand the intrinsic characteristics between modalities, thereby yielding its complete modeling capabilities and optimal information utilization. To realize this order, an asymmetric design of the model is necessary.

To further explain the advantages of the asymmetric design above, four modeling strategies are illustrated in Figure \ref{fig: intro}, namely independent modeling, unidirectional modeling, symmetric modeling, and asymmetric modeling respectively. Independent modeling lacks interactive information exchange during the modeling process, while unidirectional modeling ignores the information supplementation from A to B, which makes it challenging to achieve modal complementarity. Besides, symmetric modeling overlooks intrinsic characteristics between two modalities, thus the modeling potential may not be fully exploited. The modeling strategy applied in our method initially utilizes information from B to guide the modeling of A, followed by using the optimized features $\mathcal{F}(A\gets B)$ to guide the modeling of B. This asymmetric design ensures comprehensive information complementarity between modalities, maximally the modeling capabilities, and the valuable information from both modalities.

According to the analysis above, a modality-sensitive and asymmetrically-designed multimodal fusion method for gait recognition is proposed in this paper, named \textbf{Li}DAR-\textbf{C}amera \textbf{A}symmetric \textbf{F}usion (\textbf{LiCAF}). This method effectively understands the intrinsic characteristics of gait depth images and silhouettes, thus it is specifically suitable for LiDAR-camera fusion. By employing temporal modeling, LiCAF accurately captures and integrates human motion information from both LiDAR and camera data, which is suitable for gait recognition. It also adopts an asymmetric modeling strategy, which initially utilizes silhouettes to guide the modeling of depth images, and then the optimized high-quality depth images guide the modeling of silhouettes. Notably, during the modeling process, LiCAF utilizes global or aggregated temporal information from one modality for the modeling of the other, thereby eliminating the need for strict temporal alignment between the two modalities.

The proposed LiCAF comprises two carefully designed serial modules: Asymmetric Cross-modal Channel Attention (\textbf{ACCA}) and Interlaced Cross-modal Temporal Modeling (\textbf{ICTM}). 
ACCA and ICTM each fulfill distinct functions, therefore considering the inherent characteristics of input features from both modalities and designing the guidance order for modeling are crucial.
ACCA is designed to enhance channels with valuable information in both modalities and suppress channels with irrelevant information, thus allowing subsequent temporal modeling to focus more on beneficial information.
Since low-quality silhouettes with plenty of noise need to be modeled under the guidance of useful information from both modalities, the asymmetric design of ACCA first utilizes silhouettes for guiding the modeling of depth images, then the optimized high-quality depth images guide the modeling of silhouettes.
This design makes ACCA minimize the noise in silhouettes, reaching its full potential.
Moreover, ICTM utilizes the silhouette and depth image features after ACCA, which are more useful and gait-related for temporal modeling. It initially leverages global temporal information from the depth images, which is more valuable, to guide the temporal modeling of silhouettes. Subsequently, the optimized latter is used to guide and refine the former. Such an asymmetric design ensures adequate information interaction between the two modalities during robust temporal modeling, which is more suitable for LiDAR-camera fusion in gait recognition.

In summary, our contributions are as follows: \textbf{(1)} We propose a novel modality-sensitive multimodal gait recognition framework that utilizes an asymmetric design mechanism to focus on different intrinsic characteristics of two modalities, \textit{i.e.} silhouettes and depth images. \textbf{(2)} Specifically, two key modules with asymmetric structure are proposed to achieve the valuable channel information selection and cross-modal temporal modeling, respectively.
\textbf{(3)} Extensive experiments are conducted on the SUSTech1K dataset, validating the powerful and robust capabilities of our LiCAF framework, and demonstrating its state-of-the-art performance. Additionally, numerous ablation experiments are designed to confirm the effectiveness of each module.

\begin{figure*}[t]
    \centering
    \includegraphics[width=1\linewidth]{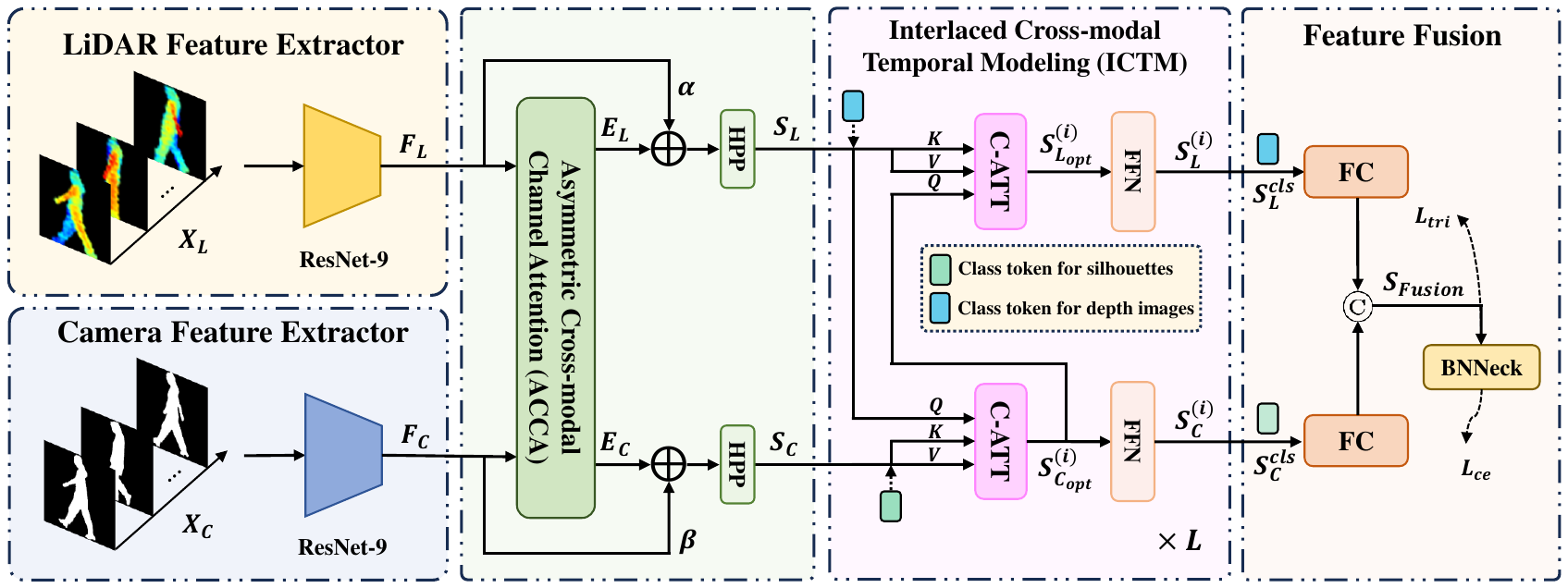}
    \caption{Overview of LiCAF. Depth image features $F_L$ and silhouette features $F_C$ are obtained from the LiDAR Feature Extractor and Camera Feature Extractor. ACCA selects channel information from $F_L$ and $F_C$, resulting in channel-enhanced features $E_L$ and $E_C$. Subsequently, $E_L$ and $E_C$ are passed through an HPP operation to form the inputs $S_L$ and $S_C$ for ICTM. Next, ICTM performs temporal modeling with $L$ layers, yielding camera features $S_L^{cls}$ and LiDAR features $S_C^{cls}$. Finally, the fusion of $S_L^{cls}$ and $S_C^{cls}$ results in the gait representation $S_{Fusion}$. Here, $L_{tri}$ and $L_{ce}$ represent triplet loss and cross-entropy loss respectively, and $\alpha$ and $\beta$ after ACCA are learnable weights.}
    \label{fig: overview}
    \vspace{-0.2cm}
\end{figure*}

\section{Related Work}
\label{sec: Related Work}

\textbf{Gait Recognition.} Unimodal gait recognition methods can be broadly categorized into two types: appearance-based methods and model-based methods. The former methods focus on the visual aspects of human movement, using features extracted directly from images or videos \cite{lin2021gait, huang2021context, huang20213d, rao2023multi}, while the latter methods analyze human structure to obtain poses or skeletons, thereby extracting gait-related features \cite{zhang2023spatial, huang2023condition, pinyoanuntapong2023gaitmixer, teepe2022towards}. In recent years, multi-modal fusion in gait recognition has achieved significant advancements, addressing some limitations in unimodal gait recognition methods. Zheng \textit{et al.} \cite{zheng2022gait} proposed SMPLGait, which fuses 2D features extracted from silhouettes with knowledge of 3D viewpoints and shapes learned from the 3D-SMPL model. Peng \textit{et al.} \cite{peng2023learning} introduced a network named BiFusion, which explores discriminative gait patterns in skeletons, integrating them with silhouette representations to learn rich features. Cui \textit{et al.} \cite{cui2023multi} designed a multimodal fusion gait recognition framework called MMGaitFormer, which utilizes cross-attention to achieve both temporal and spatial multimodal fusion, resulting a comprehensive gait representations from silhouettes and skeletons.

\textbf{LiDAR-camera Fusion.} Recent studies mainly utilize multimodal fusion networks to simultaneously take advantage of both LiDAR and camera data, which achieve significant advancements in fields such as 3D detection, object recognition, and map reconstruction \cite{bai2022transfusion, li2022bevformer, wang2023lidar2map}. 
Liang \textit{et al.} \cite{liang2022bevfusion} proposed a simple yet effective framework called BEVFusion, which models BEV features by point cloud and image streams, employing a Dynamic Fusion Module to fuse camera and LiDAR BEV features.
Wang \textit{et al.} \cite{wang2023lidar2map} designed an efficient framework named LiDAR2Map for semantic map construction, which applies an effective camera-to-LiDAR distillation scheme, ensuring that LiDAR features can thoroughly absorb the semantic information from images.
However, most methods, which employ a coarse-grained fusing approach and ignore temporal modeling, are unsuitable for gait recognition. Compared to LiDAR-camera fusion in other fields, the fusion for gait recognition is currently in the initial stage of research.
Han \textit{et al.} \cite{han2022licamgait} designed the first LiDAR-camera fusion method in gait recognition, namely LiCamGait. It employs a cross-attention mechanism to fuse gait point clouds and silhouettes, leveraging depth information from the former and dense representations from the latter to achieve outstanding results. 
Shen \textit{et al.} \cite{shen2023lidargait} introduced SUSTech1K, the first large-scale LiDAR-camera-based dataset in gait recognition, which encompasses various challenging walking conditions.

Consequently, to achieve LiDAR-camera fusion based on fine-grained fusion and temporal modeling of gait information, we propose a modality-sensitive method named LiCAF, which understands the intrinsic characteristics of LiDAR and camera data. It employs an asymmetric design to ensure comprehensive information interaction between modalities and utilizes fine-grained temporal modeling to acquire robust gait representations, which is one of the forefront methods for LiDAR-camera fusion in gait recognition.

\section{Method}
\label{sec: Method}

In this section, we present the overview of the proposed LiCAF pipeline and the details of its key modules, namely the ACCA and ICTM modules. We also introduce the loss functions employed in this method. 

\subsection{Pipeline}
\label{ssec:Pipeline}

Figure \ref{fig: overview} illustrates our LiDAR-camera multimodal fusion framework, LiCAF, which encompasses two primary and cascaded components: ACCA and ICTM. 
The network accepts inputs of depth image sequence $X_L$ and silhouette sequence $X_C$, with dimensions $3 \times T_L \times H \times W$ and $1 \times T_C \times H \times W$, respectively. 
Initially, ResNet-9 is employed as both the camera feature extractor and LiDAR feature extractor, thus the feature maps of depth images $F_L$ and silhouettes $F_C$ are obtained. Subsequently, the ACCA module enhances useful channels and suppresses irrelevant channels in both modalities, yielding channel-enhanced features $E_L$ and $E_C$. 
Before temporal modeling, the Horizontal Pyramid Pooling (HPP) operation is applied to horizontally partition the feature maps into $P$ parts, resulting in:
\begin{equation}
    \begin{cases}
    S_L = \mathrm{HPP}(\alpha F_L + E_L)\in \mathbb{R}^{C_1 \times T_L \times P}
    \\
    S_C = \mathrm{HPP}(\beta F_C + E_C)\in \mathbb{R}^{C_1 \times T_C \times P} 
    \end{cases},
\end{equation}
where $\alpha$ and $\beta$ are learnable weights. Next, the ICTM employs $L$ layers, each with two cascaded Transformer blocks, for temporal modeling. In the $i$-th layer, LiDAR features first serve as query (Q) to guide the temporal modeling of camera features, resulting in the optimized features $S_{C_{opt}}^{(i)}$. Then $S_{C_{opt}}^{(i)}$ acts as Q to further guide the temporal modeling of LiDAR features, yielding the modeled LiDAR features $S_{L_{opt}}^{(i)}$. Then, these two optimized features are followed by Feed Forward Networks (FFNs). Finally, the class tokens of both modalities, $S_L^{cls}$ and $S_C^{cls}$, are obtained, passed through an FC layer, and then concatenated to form the final gait representations $S_{Fusion}$. The network is trained by using a combined loss function of triplet loss and cross-entropy loss.

\subsection{Asymmetric Cross-modal Channel Attention (ACCA)}
\label{ssec:GCCA}

\begin{figure}[t]
    \centering
    \includegraphics[width=0.8\linewidth]{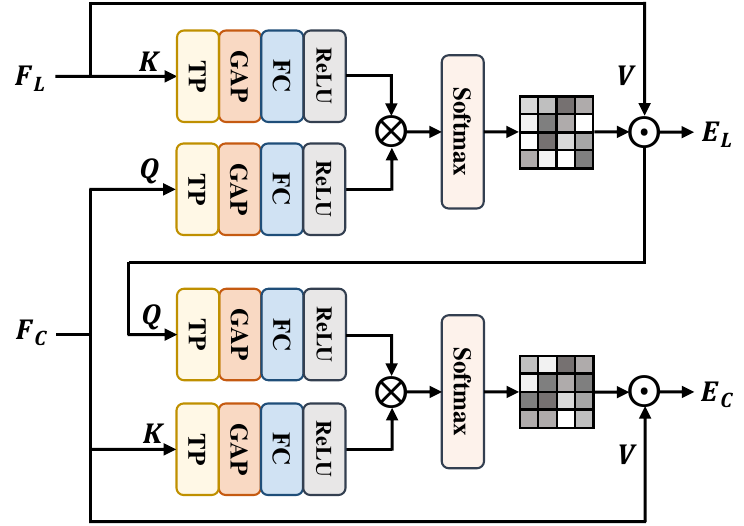}
    \caption{The detailed structure of ACCA. Here, $\mathrm{TP}$ denotes temporal max pooling, $\mathrm{ GAP }$ represents spatial global average pooling, $\mathrm{ FC }$ refers to a linear projection layer, and $\mathrm{ ReLU }$ indicates the activation function.}
    \label{fig: acca}
    \vspace{-0.2cm}
\end{figure}

We propose an asymmetric channel attention mechanism for cross-modal guidance in channel information selection, with its structure depicted in Figure \ref{fig: acca}. 
Specifically, LiDAR features tend to carry more representative gait features than camera features do, and camera features often necessity useful information from both modalities for modeling due to plenty of noise and less gait information. Consequently, the ACCA module first uses silhouettes to guide depth images, enhancing the useful information and suppressing the irrelevant information. 
During this process, $F_L \in \mathbb{R}^{C_0 \times T_L \times H \times W}$ and $F_C \in \mathbb{R}^{C_0 \times T_C \times H \times W}$ first undergo temporal-spatial integration denoted as $\Gamma(\cdot)$, then followed by calculating cross-modal channel attention map and finally weighing the depth images. The entire process can be expressed as: 
\begin{equation}
    E_L = \mathrm{Softmax}(\Gamma (F_C) \otimes \Gamma (F_L)^T) F_L
\end{equation}
where $\Gamma(\cdot) = \mathrm{ReLU}(\mathrm{FC}(\mathrm{GAP}(\mathrm{TP}(\cdot))))$.
Hence, the channel-enhanced depth image features $E_L$ are obtained. Afterward, we utilize $E_L$ to guide the selection of channel information in silhouette features, then the channel-enhanced silhouette features $E_C$ are derived: 
\begin{equation}
    E_C = \mathrm{Softmax}(\Gamma (E_L) \otimes \Gamma (F_C)^T) F_C
\end{equation}
The channel-enhanced features $E_L$ and $E_C$ enable the subsequent modeling process to focus more on valuable information. Moreover, this asymmetric channel information selection mechanism aligns with the inherent characteristics between depth image features and silhouette features, thereby maximizing the effectiveness of this mechanism. 

\begin{table*}[htbp]
    \footnotesize
    \centering
    \caption{Average Rank-1 and Rank-5 accuracy (\%) on SUSTech1K test set. The bold and underlined values represent the best and the second-best results respectively.}
    \begin{tabular}{cc|cccccccc|cc}
        \toprule
        \multirow{2}{*}{Model} & \multirow{2}{*}{Modality} & \multicolumn{8}{c|}{Probe Sequence (Rank-1 accuracy \%)}  & \multicolumn{2}{c}{Overall Acc(\%)} \\ &  & Normal & Bag & Clothing & Carrying & Umbrella & Uniform & Occlusion & Night & Rank-1 & Rank-5 \\ \hline
        
        GaitGL \cite{lin2022gaitgl} & \multirow{2}{*}{Camera} & 67.2 & 66.2 & 35.6 & 63.3 & 61.6 & 58.1 & 66.6 & 17.9 & 63.0 & 82.8 \\
        GaitBase \cite{fan2023opengait}  &  & 81.3 & 77.3 & 49.6 & 75.7 & 75.4 & 76.7 & 81.4 & 25.8 & 76.0 & 89.1 \\ \hline
        SimpleView \cite{goyal2021revisiting} & \multirow{2}{*}{LiDAR} & 72.6 & 68.9 & 57.2 & 63.3 & 49.2 & 62.5 & 79.7 & 66.5 & 65.0 & 86.0 \\
        LidarGait \cite{shen2023lidargait}  &  & 91.8 & 88.6 & 74.6 & 89.0 & 67.5 & 80.9 & 94.5 & \textbf{90.4} & 86.8 & 96.1 \\ \hline
        MMGaitFormer \cite{cui2023multi} & \multirow{5}{*}{LiDAR-Camera} & \underline{94.3} & \underline{93.7} & 80.0 & \underline{91.8} & \underline{84.0} & 88.7 & \underline{95.7} & 86.0 & \underline{91.1} & \underline{98.2} \\
        LiCamGait \cite{han2022licamgait}   &  & 84.0 & 83.5 & 67.1 & 80.1 & 69.2 & 72.9 & 89.2 & 79.5 & 80.4 & 94.9 \\
        CMOT \cite{mukhtar2024cmot}  &  & 94.1 & 92.5 & \underline{80.9} & 91.7 & 79.6 & 87.0 & 95.6 & 87.5 & 90.4 & 97.5 \\
        \textbf{LiCAF(Ours)}  &  & \textbf{95.8} & \textbf{95.7} & \textbf{82.7} & \textbf{94.5} & \textbf{89.3} & \textbf{93.6} & \textbf{96.6} & \underline{88.7} & \textbf{93.9} & \textbf{98.8} \\
        \bottomrule
\end{tabular}
    \label{tab: sota}
    \vspace{-0.5cm}
\end{table*}

\subsection{Interlaced Cross-modal Temporal Modeling (ICTM)}
\label{ssec:ICTM}

To better achieve cross-modal temporal modeling based on the information complementarity of LiDAR and camera features, ICTM is also designed in an asymmetric pattern, with its structural details displayed in Figure \ref{fig: overview}. 
Adopting the temporal integration strategy from ViT \cite{dosovitskiy2020image}, we concatenate learnable class tokens $S_L^{cls}$ and $S_C^{cls}$ to the depth image features $S_L$ and silhouette features $S_C$, respectively, forming the input for the first layer: 
\begin{equation}
    \begin{cases}
    S_L^{(0)}=\mathrm{Concat}(S_L^{cls}, S_L) \in \mathbb{R}^{C_1 \times (T_L+1) \times P}
    \\
    S_C^{(0)}=\mathrm{Concat}(S_C^{cls}, S_C) \in \mathbb{R}^{C_1 \times (T_C+1) \times P}
    \end{cases},
\end{equation}
where $\mathrm{Concat}$ is matrix concatenation. 
For ICTM, both inputs $S_L$ and $S_C$ contain more valuable information and less noise after ACCA, therefore ICTM applies a different modeling strategy compared to ACCA.
In each layer, ICTM first utilizes global temporal information of depth image features to guide the temporal modeling of silhouette features, resulting in optimized features $S_{C_{opt}}^{(i)}$. Next, $S_{C_{opt}}^{(i)}$ is used to guide the temporal modeling of the depth image features. This process can be represented as: 
\begin{equation}
    \begin{cases}
    S_{C_{opt}}^{(i)}=(\mathrm{\mathrm{C \text{-} ATT}}(S_L^{(i-1)}), S_C^{(i-1)}, S_C^{(i-1)}))
    \\
    S_{L_{opt}}^{(i)}=(\mathrm{\mathrm{C \text{-} ATT}}(S_{C_{opt}}^{(i)}, S_L^{(i-1)}, S_L^{(i-1)})
    \\
    S_L^{(i)}=\mathrm{FFN}(S_{L_{opt}}^{(i)})
    \quad \text{and} \quad
    S_C^{(i)}=\mathrm{FFN}(S_{C_{opt}}^{(i)})
    \end{cases},
\end{equation}
where $i$ denotes the $i$-th layer ($i=1,2,...,L$), $\mathrm{FFN}(\cdot)$ represents the feed forward network. Besides, $\mathrm{C \text{-} ATT}(Q, K, V)$ here represents the multi-head cross-attention mechanism, where $Q$, $K$ and $V$ denote query, key and value respectively. 

ICTM leverages the powerful modeling capabilities of transformers, utilizing the cross-attention mechanism to ensure that the temporal modeling information for a single frame of one modality is derived from all frames of another modality. This not only ensures the effective utilization of cross-modal temporal information but also guarantees comprehensive information interaction between the two modalities during the modeling process.

Finally, the class tokens $S_L^{cls}  \in \mathbb{R}^{C_2 \times P}$ and $S_C^{cls} \in \mathbb{R}^{C_2 \times P}$ in $S_L^{(L)}$ and $S_C^{(L)}$ respectively are taken as the output of ICTM, which integrate all the temporal information. After ICTM, LiDAR and camera features are fused, then fusion features can be obtained:
\begin{equation}
    S_{Fusion} = \mathrm{Concat}(\mathrm{FC}(S_L^{cls}), \mathrm{FC}(S_C^{cls})) \in \mathbb{R}^{C_3 \times P}
\end{equation}

\subsection{Loss Function}
\label{ssec:Loss}

The model is trained by using a weighted combination of triplet loss $L_{tri}$ and cross-entropy loss $L_{ce}$: 
\begin{equation}
    L=L_{tri}+L_{ce}
    \label{eq:loss}
\end{equation}
The role of the triplet loss is to maximize the inter-class distance and minimize the intra-class distance, while the cross-entropy function aims to enhance the accuracy of classifying different IDs.

\section{Experiments}
\label{sec: Experiments}

\subsection{Dataset}
\label{ssec:Dataset}

\textbf{SUSTech1K.} SUSTech1K \cite{shen2023lidargait} is the first and only large-scale public LiDAR-camera-based multimodal gait recognition benchmark, collected by industrial cameras and 128-beam LiDAR sensors. This dataset includes 1,050 identities with 25,239 sequences, where subjects adhered to specific walking rules. It provides temporally synchronized RGB and point cloud streams captured at 30 FPS and 10 FPS respectively, and encompasses various challenging walking conditions like clothing and occlusion. All experiments were conducted on SUSTech1K, which was divided into a training set (250 IDs) and a testing set (850 IDs), with grouping sequences under normal conditions into gallery sets and those under variant conditions into probe sets.

\subsection{Implementation Details}
\label{ssec:Implementation}

\textbf{Dataset Pretreatment.} We employ the depth image generation strategy provided by LidarGait \cite{shen2023lidargait} to convert point clouds into depth images. Moreover, we retain the silhouettes in pretreatment, ensuring that the ratio of silhouettes and depth images maintains 3:1, which can maximize the utilization of gait information in the dataset.

\textbf{Hyper-parameters.} For the camera and LiDAR feature extractors, LiCAF employs ResNet-9, with an output channel of 512. The HPP operation horizontally partitions the silhouettes and depth images into $[1, 2, 4, 8, 16]$ parts and then concatenates them, resulting in a total of $P=31$ parts. The number of layers $L$ in ICTM is set to 2, and the number of heads in the C-ATT block is set to 16. For the feature maps output by ACCA and ICTM from the two modalities, the channel is set to 512, which means $C_0=C_1=C_2=512$. After passing through the FC layer, the channel of the two feature maps is reduced to 128. Finally, they are concatenated to obtain the fused gait representation, with $C_3=256$.

\textbf{Training Details.} During the training phase, we apply the Batch All sampling strategy. The batch size $(p, k)$ is set to $(8, 8)$, where $p$ represents the number of IDs, and $k$ denotes the number of training sample sequences per ID. Both silhouettes and depth images have a resolution of $64 \times 64$, and their sequence frame lengths are set to 21 and 7, respectively. The margin for the triplet loss function is set to 0.2. SGD is used as the optimizer with a weight decay of $5 \times 10^{-4}$ and an initial learning rate of 0.1. The learning rate is decreased to $\times 0.1$ at 20K and 30K iterations, with the total number of training iterations set to 40K. During the testing phase, entire sequences of silhouettes and depth images are fed into the network to extract the gait representations.

\subsection{Comparison with State-of-the-art Methods}
\label{ssec:SOTA}

As presented in Table \ref{tab: sota}, our method achieves Rank-1 and Rank-5 accuracy of 93.9\% and 98.8\% on SUSTech1K respectively, and outperforms the current best-performing LidarGait \cite{shen2023lidargait} by 7.1\% in Rank-1. 
Compared to unimodal approaches, LiCAF significantly improves overall accuracy and performance under all walking conditions but night, demonstrating the effectiveness of LiDAR-camera fusion. 
For night condition, we find that camera-only methods such as GaitGL \cite{lin2022gaitgl} and GaitBase \cite{fan2023opengait} perform extremely poorly, which is 45.1\%/50.2\% lower than the overall condition. This indicates that the quality of silhouettes at night is too poor, which compromises network performance significantly. Despite a minor 1.7\% decrease compared to LidarGait at night, LiCAF improves overall accuracy by 7.1\%, outperforming other fusion methods. These results demonstrate LiCAF’s strong capability to extract and fuse useful information between modalities.
Moreover, we retrained MMGaitFormer \cite{cui2023multi}, CMOT \cite{mukhtar2024cmot}, and LiCamGait \cite{han2022licamgait} on SUSTech1K and achieved outstanding performance. Then we compared LiCAF with these methods, outperforming them by 2.8\%, 3.5\%, and 13.5\% in Rank-1 respectively, and achieving the best results under all conditions.
These impressive experimental results stem from LiCAF's ability to grasp the inherent characteristics of modalities and its asymmetric design that maximizes modeling potential, effectively utilizing beneficial information from both modalities. LiCAF's modality-sensitivity enhances its suitability for fusing silhouettes and depth images in gait recognition, showcasing its superiority.

\subsection{Ablation Study}
\label{ssec:Ablation}

\textbf{Effectiveness of main modules.} As shown in Table \ref{tab: module effectiveness}, we design ablation experiments to validate the effectiveness of the two main modules of LiCAF, namely ACCA and ICTM. The results reveal that the application of ICTM contributes a 1.2\% improvement in Rank-1 accuracy compared to the baseline, highlighting its robust cross-modal temporal information modeling capability. Additionally, ACCA provides an extra 0.5\% performance enhancement, proving the effectiveness of its cross-modal channel valuable information selection.

\begin{table}[h]
    \vspace{-0.2cm}
    \centering
    \caption{Study of the effectiveness of main components in LiCAF in terms of average Rank-1 and Rank-5 accuracy (\%) on SUSTech1K.}
    \begin{tabular}{ccc|cc}
    \toprule
    \multirow{2}{*}{Baseline} & \multirow{2}{*}{ICTM} & \multirow{2}{*}{ACCA} & \multicolumn{2}{c}{Overall Acc(\%)} \\
        &  &  & Rank-1 & Rank-5 \\ \hline
    \checkmark &   &  & 92.2 & 98.3 \\
    \checkmark & \checkmark  &  & 93.4  & 98.7 \\
    \checkmark  & \checkmark  & \checkmark  & \textbf{93.9} & \textbf{98.8}  \\ \bottomrule
    \end{tabular}
    \label{tab: module effectiveness}
    \vspace{-0.2cm}
\end{table}

\textbf{Modeling strategy in ICTM and ACCA.} Table \ref{tab: Modeling strategy in ICTM} and \ref{tab: Modeling strategy in ACCA} respectively show various modeling strategies of ICTM and ACCA, reflecting the effectiveness of the asymmetric design of these two modules, demonstrating that our strategies in both modules are optimal. 
Taking ICTM as an example for analysis, if the modeling process only involves depth images supplementing silhouette, or vice versa, the information interaction between modalities during modeling is insufficient. As a result, these two strategies achieve the performance of 87.1\% and 87.0\% respectively. However, considering both information supplementing simultaneously can reach the performance of 93.4\%, with an increase of 6.3\% and 6.4\% over the former two strategies respectively. Similar results are also observed in ACCA (Table \ref{tab: Modeling strategy in ACCA}). 
Moreover, results of ACCA in Table \ref{tab: Modeling strategy in ACCA} show that the strategy of first utilizing silhouettes for guiding the channel selection of depth images and then vice versa is superior, surpassing the reverse order and simultaneous modeling strategies by 0.5\% and 0.1\%. 
The reason for this is that the silhouettes contain a significant amount of noise, necessitating modeling guidance from the information of both modalities.
From Table \ref{tab: Modeling strategy in ICTM}, we also find that our strategy in ICTM, where depth images guide silhouette temporal modeling first and then vice versa, outperforms the reverse order and simultaneous modeling strategies, with improvements of 0.3\% and 0.5\%, respectively. This is because depth images and silhouettes contain less gait-irrelevant information after ACCA, and the former often carry more useful temporal information than silhouettes do, making it a wiser strategy to first use depth images to enhance the quality of silhouette modeling. In general, the experimental results demonstrate the modality-sensitivity of the two proposed modules, highlighting the importance of the order in information supplementation and modeling guidance between two modalities.

\begin{table}[h]
    \vspace{-0.3cm}
    \small
    \centering
    \caption{Comparison of modeling strategy applied in ICTM in terms of average Rank-1 and Rank-5 accuracy (\%) on SUSTech1K. Here, $C$ and $L$ respectively represent the silhouettes and depth images. $C\gets L$ denotes the supplementation of information from depth images during the modeling of silhouettes, while $\mathcal{F}(C\gets L)$ represents the result of this modeling process.}
    \begin{tabular}{c|cc}
        \toprule
        \multirow{2}{*}{Modeling strategy in ICTM} & \multicolumn{2}{c}{Overall Acc(\%)} \\
        & Rank-1 & Rank-5 \\ \hline
        $C\gets L$ only & 87.1  & 96.8 \\
        $L\gets C$ only & 87.0  & 96.6 \\
        $C\gets L$ and $L\gets C$ simultaneously   & 93.4  & 98.6  \\
        $L\gets C$, then $C\gets \mathcal{F}(L\gets C)$          & 93.6  & 98.7 \\
        $C\gets L$, then $L\gets \mathcal{F}(C\gets L)$ \textbf{(ours)}    & \textbf{93.9}  & \textbf{98.8 } \\ \bottomrule
    \end{tabular}
    \label{tab: Modeling strategy in ICTM}
    \vspace{-0.7cm}
\end{table}

\begin{table}[h]
    \small
    \centering
    \caption{Comparison of modeling strategy applied in ACCA in terms of average Rank-1 and Rank-5 accuracy (\%) on SUSTech1K.}
    \begin{tabular}{c|cc}
        \toprule
        \multirow{2}{*}{Modeling strategy in ACCA} & \multicolumn{2}{c}{Overall Acc(\%)} \\
        & Rank-1 & Rank-5 \\ \hline
        $C\gets L$ only & 93.7  & 98.7 \\
        $L\gets C$ only & 93.7  & 98.7\\
        $C\gets L$ and $L\gets C$ simultaneously   & 93.8  & 98.7  \\
        $C\gets L$, then $L\gets \mathcal{F}(C\gets L)$         & 93.4  & 98.7 \\
        $L\gets C$, then $C\gets \mathcal{F}(L\gets C)$  \textbf{(ours)}   & \textbf{93.9}  & \textbf{98.8}  \\ 
        \bottomrule
    \end{tabular}
    \label{tab: Modeling strategy in ACCA}
    \vspace{-0.3cm}
\end{table}

\section{Conclusion}
\label{sec: Conclusion}

In this paper, we introduce LiCAF, a modality-sensitive gait recognition method for LiDAR-camera fusion. LiCAF understands the inherent characteristics of silhouettes and depth images, and adopts an asymmetric modeling strategy optimal for their fusion. LiCAF achieves state-of-the-art performance on SUSTech1K, with extensive experiments validating the effectiveness of its components. As LiDAR-camera fusion for gait recognition is in the initial stage, we aim to further explore new methods in this field in the future.



\footnotesize
\bibliographystyle{IEEEbib}
\bibliography{refs}

\end{document}